# Equitable and Fair Performance Evaluation of Whale Optimization Algorithm

Bryar A. Hassan[1], Tarik A. Rashid[2], Aram Ahmed[1,3], Shko M. Qader[4,5], Jaffer Majidpour[6], Mohmad Hussein Abdalla[6], Noor Tayfor[7], Hozan K. Hamarashid[8], Haval Sidqi[9], Kaniaw A. Noori[10], Awf Abdulrahmam Ramadhan[11]

[1]Department of Computer Science, College of Science, Charmo University, Chamchamal 46023, Sulaimani, KRI, Iraq

[2]Department of Computer Science and Engineering, University of Kurdistan Hewler, KRI, F. R Iraq.

[3]Department of Information Technology, College of Science and Technology, University of Human Development, Sulaimani, Iraq

[4]Computer Science Institute, Sulaimani Polytechnic University, Information Technology Department, Sulaimani, Iraq

[6]The Department of Computer Science, Kurdistan Technical Institute, Sulaimani, Iraq

[6]Department of Computer Science, University of Raparin, Sulaimani, Iraq

[7]Department of Information Technology, Kurdistan Technical Institute, Sulaimani, Iraq

[8]Information Technology Department, Computer Science Institute, Sulaimani Polytechnic University, Sulaimani 46001, Iraq

[9]Technical College of informatics, Sulaimani Polytechnic University, Sulaimani 46001, Iraq

[10]Department of Database Technology, Technical College of Informatics, Sulaimani Polytechnic University, Sulaimani, Iraq

[11]Public Health Department, College of Health and Medical Techniques - Shekhan, Duhok Polytechnic University, Duhok, Iraq

Email (corresponding): bryar.hassan@charmouniversity.org; tarik.rashid@ukh.edu.krd

## Abstract

It is essential that all algorithms are exhaustively, somewhat, and intelligently evaluated. Nonetheless, evaluating the effectiveness of optimization algorithms equitably and fairly is not an easy process for various reasons. Choosing and initializing essential parameters, such as the size issues of the search area for each method and the number of iterations required to reduce the issues, might be particularly challenging. As a result, this chapter aims to contrast the Whale Optimization Algorithm (WOA) with the most recent algorithms on a selected set of benchmark problems with varying benchmark function hardness scores and initial control parameters comparable problem dimensions and search space. When solving a wide range of numerical optimization problems with varying difficulty scores, dimensions, and search areas, the experimental findings suggest that WOA may be statistically superior or inferior to the preceding algorithms referencing convergence speed, running time, and memory utilization.

## 1. Introduction

In a rapidly growing field such as optimization algorithms, all algorithms must be assessed accurately, equitably, and intelligently. This endeavor is not simple for various reasons. Due to the random nature of the processes involved, it is necessary to do several test runs of the algorithms to average out the effects of this unpredictability. Numerous published tests of algorithms provide a mean result, standard deviation, and the best and worst outcomes to illustrate the performance characteristics of the algorithm. When comparing several algorithms' performance, we face several problems. Among these are the utilized computer code's efficacy, the algorithms' unpredictability, and setting parameters at an equitable and similar level. This final issue deserves clarification. Consider that we wish to compare the firefly method to the simulated annealing technique. The firefly algorithm must provide the number of fireflies, the parameters, and 0. How can we choose these characteristics to compare the two approaches appropriately? A wrong choice of quenching factor or starting temperature for the particular test issue would result in poor performance of the simulated annealing methodology, preventing a fair comparison of the techniques. Comparing methods is not difficult, but prudence is required. In addition, specific algorithms may perform very well on one problem but poorly on another or minor problems but poorly on large ones. Therefore, a comprehensive collection of standard test problems that reflect the evolving difficulties of nonlinear optimization must be used. Statistical significance tests like Friedman's may be helpful in comparative research. See Daniel for further information on Friedman's experiment (1990). In addition, the tests must be done with identical settings, and efficiency assessments can be determined from time measurements or function evaluations. Function assessments provide a more precise comparison of testing in diverse computer settings. The evaluation test tasks for nature-inspired algorithms are vital. These should test the algorithm's fundamental properties. The essential property is that algorithms can calculate the global optimum of a function. Therefore, problems with several local optima must be included in the test set. Moreover, the distribution of these optimums is a crucial feature. Because both types of functions must be optimized and provide separate problems, specific test questions should yield optimum, tightly-packed answers. In contrast, others should generate ideal solutions that are widely spread and isolated. A sharp, practically discontinuous approach to the ideal valley or peak, or a steady descent to an undefined minimum, must also be considered. Fortunately, a rigorous and well-respected set of test problems reflecting these and other properties has been compiled, and new algorithms are routinely assessed on this set. The reader may be surprised to learn that accurate results may not be achieved during these tests. Some problems are intended to be pathologically tricky; hence, only highly approximate values for the optimum may be established. The search will be affected by the size or number of parameters of the problem. In such cases, we seek the algorithm that produces the most outstanding results, and academics in the field normalize the data to reflect this. Since this is not a collection of research papers, we refrained from conducting tests on monumental topics. Smaller jobs allow us to demonstrate the algorithm's essential properties so the reader can duplicate reasonably, even on a simple computer.

## 2. Background

Evaluation of the effectiveness of optimization algorithms is an intriguing area of study. However, several obstacles exist to conducting a fair and objective performance analysis of these algorithms. These aspects may include selecting initial settings for competing algorithms, system performance during the evaluation, algorithmic programming style, the proportion of randomization, and the cohorts and difficulty ratings of the chosen benchmark problems. In this chapter, we will choose the most up-to-date and well-known optimization strategies for equal evaluation versus WOA because the literature lacks equal experimental evaluation research on WOA. The selected algorithms are the Backtracking Search Optimization Method (BSA), Fitness Dependent Optimizer (FDO), Particle Swarm Optimization Algorithm (PSO), and Firefly Algorithm (FF). In this part, the prior algorithms involving WOA are briefly described.

### 2.1. WOA

WOA is a suggested swarm intelligence method for continuous optimization issues. This method has been demonstrated to perform as well as or better than specific existing algorithmic strategies [1]. WOA has drawn inspiration from the hunting behavior of humpback whales. In WOA, each solution is referred to as a whale. Using the best member of the group as a reference, a whale seeks to populate a new spot in the search space. Whales have two methods for locating prey and pursuing it. In the first, prey is encircled, and bubble nets are produced in the second. Exploration of the search space occurs during the whales' hunt for prey, whereas exploitation happens during their attack behavior. The complete description and source codes of the WOA are available in [1].

## 2.2. BSA

BSA is an iterative population-based meta-heuristic algorithm. BSA creates test populations to regulate the amplitude of the search-direction matrix, providing a robust capacity for global exploration [2]. During the crossover phase, BSA utilizes two random crossover procedures to interchange the corresponding elements of people in populations and test populations. In addition, the BSA has two selection methods. One is used to pick populations from the present and historical populations, while the other determines the best population. BSA may be broken down into five distinct processes: (a) Initiation, (b) Selection I, (c) Mutation, (d) Crossover, and (e) Selection II. The BSA's complete description and source codes are available in [2].

## 2.3. FDO

FDO is comparable to a swarm of bees reproducing. This method is based on how scout bees navigate several prospective colonies, searching for a new, appropriate hive. Each scout bee that looks for additional colonies offers a potential solution to this technique. The best hive among numerous good hives must be chosen to achieve convergence to optimality [3]. Starting with a random initiation of a synthetic scout population in the search space Xi(i=1,2,...,n), each scout bee position represents a newly discovered hive. The process then repeatedly looks for bee hives (solution). Scout bees randomly investigate various locations to locate superior hives; once a superior hive is located, the previously located hive is disregarded. In the same way, the algorithm discards the previously recognized alternative when it discovers a new, superior one. The artificial scout bee might not find a better option if the current activity continues (hive). In such an instance, it will carry on along its introductory course to arrive at a better answer. Nonetheless, the algorithm will go on to the present solution, which is the best choice so far, if the prior route does not result in a better solution. The FDO's complete description and source codes may be found in [3].

## 2.4. PSO

PSO is a self-adaptive, stochastic, population-based optimization method. The PSO initially creates the first particles and gives them starting velocities [4]. The ideal function value and position are calculated by assessing the objective function at each particle size. It selects new velocities based on the present velocity, the particles' ideal positions, and their neighbors' ideal positions. The particle positions, velocities, and neighbors are then regularly updated (the new location is the previous one, plus the velocity, tweaked to keep particles inside boundaries). Up until a halting condition is encountered, the algorithm iterates. PSO's complete source code and description are available in [4].

## 2.5. FF

FF is one of the optimization algorithms inspired by nature that Xin-She Yang devised at Cambridge in 2008 and made public by Yang (2009) [5]. A technique for optimizing functions with multiple optima was created using the behavior of firefly swarms. To encourage exploration of the solution space, it specifically used the notion that individual fireflies' brilliance attracted them together as well as a randomization component.

The release of the firefly method has led to the publication of several publications on its study, modification, and application to numerous real-world issues, as well as numerous accounts of its successful uses. The essential characteristics of this algorithm are based on the following principles created by Xin-She Yang:

1. Since fireflies are unisexual, they all have a strong attraction to one another.
2. The brightness of them affects how beautiful they are. Though perceived brightness reduces as two fireflies are separated, the less brilliant one will always be drawn to (and move towards) the brighter one.
3. It will move arbitrarily if no other fireflies are brighter than a particular firefly.

These three criteria may be used to create an optimization process where the brightness is inversely correlated with the value of the goal function. The firefly rules may be turned into algorithmic steps by creating the locations of an initial population of fireflies, calculating the objective function value for each firefly, and then putting these rules into practice over several generations. Public access to FF's source code may be found in [6].

## 3. Evaluation

The mean, standard deviation, and worst and best outcomes were calculated based on earlier studies [1], [7] that compared the performance characteristics of meta-heuristic algorithms to those of their rivals. For various reasons, comparing the effectiveness of various metaheuristic algorithms fairly and similarly is challenging. The choice and initialization of crucial variables, such as the size of the issue, the search space, and the number of iterations needed to solve the problem for each strategy, pose particular difficulties. Another reason to be concerned is that every algorithm depends on a randomization mechanism. It is required to test the algorithms several times to remove the impacts of this randomization and provide an average result. The programming language and technology used to construct the approach may also impact how well an algorithm performs in classification. All algorithms on the same system can utilize the same coding approach to prevent this impact. Selecting the issues or benchmark test functions for the algorithm evaluation is another challenging undertaking. For instance, an algorithm could work well for some categories of issues but not others.

Also, the size of a task may impact an algorithm's performance. The performance of the algorithms must be assessed using a set of typical issues or benchmark test functions with a range of complexity levels and various cohorts on a range of optimization tasks to solve the aforementioned issue. The initialization of the control parameters, the balance of the randomness in the algorithms, the computer performance used to implement the algorithms, the programming style of the algorithms, and the kinds of problems they are intended to solve are all factors that must be considered in an evaluation method to accurately compare WOA to its rival algorithms in the absence of experimental research on WOA in the literature. As a consequence, this part covers the experimental design, the list of benchmark problems utilized in the assessment, their control components, the statistical analysis of the evaluation technique, the pairwise statistical testing methods, and the evaluation outcomes.

### 3.1. Evaluation method

The evaluation method considers several factors, such as the initialization of parameters like problem dimensions, problem search space, and the number of iterations necessary to minimize the problem, the performance of the system used to implement the algorithms, the programming style of the algorithms, achieving a balance on the effect of randomization, and the use of various types of optimization problems to accurately compare WOA with BSA, PSO, FDO, and FF. This process compares WOA and its rival algorithms on 16 benchmark tasks with varying degrees of difficulty [8]–[10].

- A specific function with Nvar variables and the default search space for a population of 30 has to be optimized across several rounds. Nvars can have one of three values: 10, 30, or 60. Each algorithm is executed thirty times for each benchmark function, with 2000 iterations for Nvar values of 10, 30, and 60.
- Many iterations are necessary to minimize functions with two variables in three different solution spaces with a population of thirty. Each algorithm is run 30 times with 2000 iterations for each of the three unique ranges (R1, R2, and R3): [-5, 5], [-250, 250], and [-500, 500].
- The proportion of successful function minimization for Evaluations 1 and 2 should be determined to compare the success rate of WOA to that of its competitors.

Figure (1) displays the evaluation framework that comprises the procedures of Evaluation 1, 2, and 3.

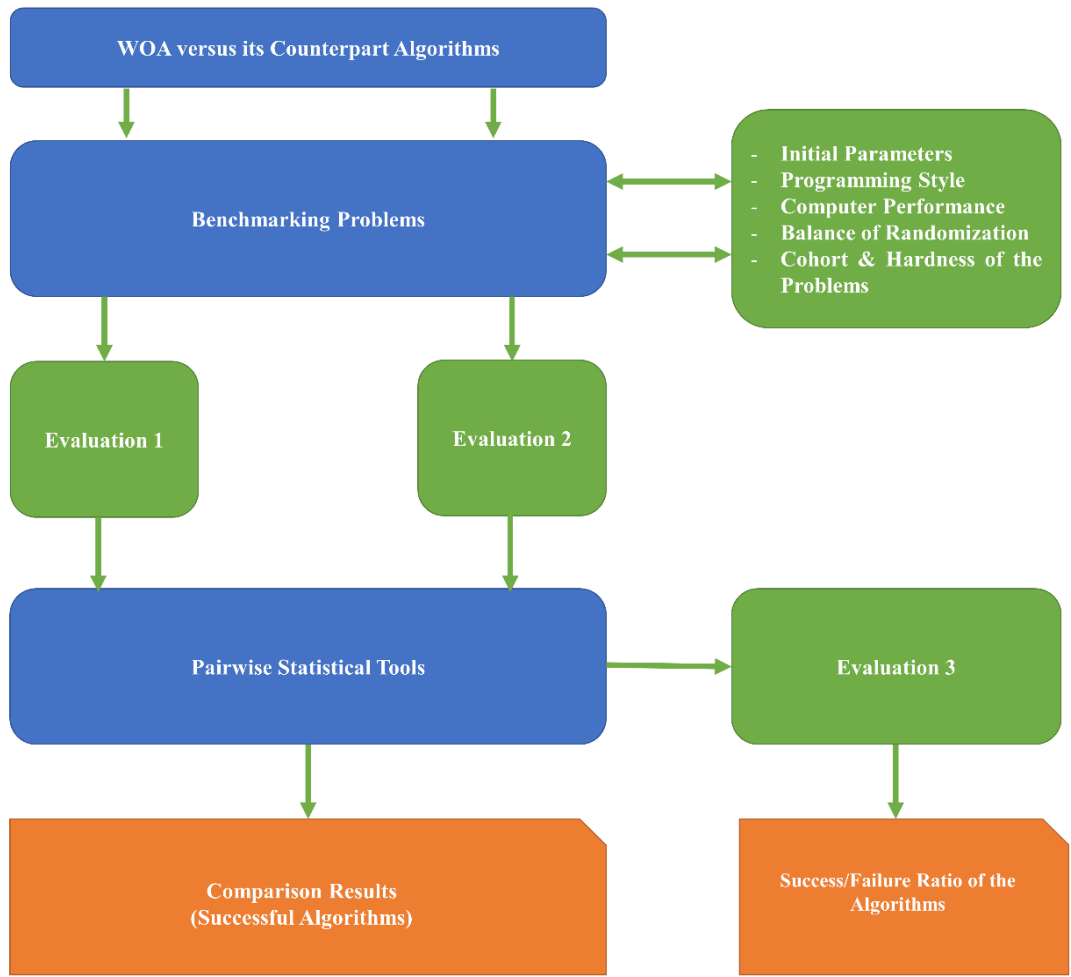

**Figure (1): The performance evaluation framework of WOA**

## 3.2. Problems and initial parameters

In each of the three assessments, 16 benchmark issues were used to evaluate the associated performance of WOA against BSA, FDO, PSO, and FF in addressing a range of optimization problems based on their cohort and complexity level [8]–[10]. Regardless of how the test function problems were resolved, specific problems are more challenging. The chosen optimization strategies, their search space, global minimum, dimension, and percentage of difficulty, are shown in Table 1. Hardness percentages can range from 4.92 to 82.75 percent. The Table's most challenging problem is the Whitley function, while its most manageable problem is the sphere.

Table (1): Problems with a different success rate

| Problem ID, Name | Global min. | Space | Dimension | Success percentage |
|---|---|---|---|---|
| P1, Ackley | 0 | [-32, 32] | n | 48.25 |
| P2, Alpine01 | 0 | [0, 10] | 2 | 65.17 |
| P3, Bird | -106.76453 | [-6.283185, | 2 | 59.00 |
| P4, Leon | 0 | [0, 10] | 2 | 41.17 |
| P5, CrossInTray | -2.062611 | [-10, 10] | 2 | 74.08 |
| P6, Easom | -1 | [-100, 100] | 2 | 26.08 |
| P7, Whitley | 0 | [-10.24, | 2 | 4.92 |
| P8, EggCrate | 0 | [-5, 5] | 2 | 64.92 |
| P9, Griewank | 0 | [-600, 600] | n | 6.08 |
| P10, HolderTable | -19.2085 | [-10, 10] | 2 | 80.08 |
| P11, Rastrigin | 0 | [-5.12, 5.12] | n | 39.50 |
| P12, Rosenbrock | 0 | [-5, 10] | n | 44.17 |
| P13, Salomon | 0 | [-100, 100] | 2 | 10.33 |
| P14, Sphere | 0 | [-1, 1] | 2 | 82.75 |
| P15, StyblinskiTang | -39.1661 | [-5, 5] | n | 70.50 |
| P16, Schwefel26 | 0 | [-500, 500] | 2 | 62.67 |

Also, for each benchmark problem, each algorithm is run thirty times with a population size of thirty for each of the three Evaluations, with a maximum of 2,000 iterations. For the benchmark tasks, Evaluations 1, 2, and 3 each have different search regions and sizes. Each benchmark problem for Evaluation 1 receives the default dimension and has its search space separated into R1, R2, and R3. Nvar1, Nvar2, and Nvar3 are the search space dimensions for each benchmark function in evaluation 2. To evaluate the success rate of WOA to that of its rival techniques, it is necessary to determine the ratio of successful function minimization for Evaluations 1 and 2. Tables (2) and (3) for Evaluations 1 and 2, respectively, illustrate the initializations of these parameters.

Table (2): Initial parameters of Evaluation 1

| Features | Dimension | Default |
|---|---|---|
| | Space | R1[-5, 5]; R2 [-250, 250] R3 [-500, 500] |
| | Hardness rate | 4.92% - 82.75% |
| Initial parameters | Number of executions | 30 |
| | iterations | 2000 |
| | Population size | 30 |

Table (3): Initial parameters of Evaluation 2

| Features | Dimension | Nvar1: 10; Nvar2: 30 and Nvar3: 60 |
|---|---|---|
| | Space | 2 |
| | Hardness rate | 4.92% - 82.75% |
| Initial parameters | Number of executions | 30 |
| | iterations | 2000 |
| | Population size | 30 |

## 3.2. Statistical analysis and tool

Occasionally, meta-heuristic algorithms might present the worst and most relevant answers for a given problem; for instance, if an algorithm runs twice on an equivalent problem, it might receive the appropriate answer the first time and the worst solution the second time, and versa. In the studies [11] and [12], statistical methods were used to contrast WOA's success or failure of WOA, problem-solving algorithms, and others. The discussed experiment resolved numerical optimization problems using seven statistical measures: mean, standard deviation, best, worst, average computing time, number of effective minimizations, and the number of failed minimizations.

To evaluate two algorithms and determine whether one has a greater statistical success rate in solving a specific optimization issue, one can use the Wilcoxon signed-rank test and other Pairwise statistical testing tools [2], [13]. The Wilcoxon signed-rank test is used in the evaluation to compare WOA to other algorithms; the statistically significant value (H0) is set at 0.05, and Equation (1) states the null hypothesis (H0) for a particular benchmark problem.

$$Median\ (Algorithm\ A) = Median\ (Algorithm\ B) \tag{1}$$

The Wilcoxon signed-rank test used R+, R-, and p-value to determine the approach that produced the statistically better response. Using GraphPad Prism, T+ and T- values ranging from 0 to 465 are determined for the same experiment. Comparable to the mathematical precision of today's application and software development tools, the P-value ranges between 4 and 6. In general, the precision used for Evaluations 1 and 2 was 6 since this level of precision may be necessary for real-world applications.

# 4. Result evaluation

This section consists of three sub-sections. The results of the three evaluations mentioned in the previous section are covered in the first sub-section. On the other hand, the computational cost of WOA compared to its comparative algorithms is described in sub-section two. Last but not least, the convergence curve of WOA is presented.

## 4.1. Results of three evaluations

Each technique decreased the four optimization issues, P3, P4, P5, and P8, in Evaluation 1. This evaluation solved optimization problems utilizing the algorithms' three dimensions and default search spaces: Nvars 1 through 3. However, since the algorithms could minimize four problems of varying dimensions, the dimensions of the numerical optimization problems did not affect the success rate of the algorithms used to minimize them. In addition, the shortened benchmark functions were of relatively low difficulty. The lowest success rate for these difficulties was 41.17 percent, while the best was 59 percent. As a result, none of the techniques effectively reduced difficulties with a high hardness score and changeable dimensions.

In contrast, problem P10, which had an overall success rate of 80.08 percent, could not be solved in any variable dimension by any approach. This indicates that a conclusive conclusion could not be made on the efficacy of WOA and other competing algorithms in reducing optimization issues based on their hardness of score. As a result, there is no intrinsic link between the difficulty of minimization optimization problems and the efficiency of these algorithms. Using problem-based statistical comparison techniques, it was established which of the experiment's algorithms could statistically resolve the benchmark functions. This method utilizes the computational time of the algorithms needed to complete 30 iterations and find the global minimum. Throughout the evaluation process, Using the Wilcoxon signed-rank test and an importance criterion () of 0.05, WOA was compared to other algorithms. Also, the Equation's null hypothesis (H0) for a specific benchmark problem was investigated (1). Tables 14, 15, and 16 show that the algorithms in Evaluation 1 statistically outperformed the other algorithms according to the Wilcoxon signed-rank test (Tables 4, 5, and 6).

**Table (4): Statistical evaluation utilizing the two-sided Wilcoxon Signed-Rank Test (α=0.05) to identify the best solution for the problems solved in Evaluation 1 (Nvar1)**

| Problems | WOA vs. BSA | | | | WOA vs. FDO | | | | WOA vs. PSO | | | | WOA vs. FF | | | |
|---|---|---|---|---|---|---|---|---|---|---|---|---|---|---|---|---|
| | p-value | T+ | T- | Winner | p-value | T+ | T- | Winner | p-value | T+ | T- | Winner | p-value | T+ | T- | Winner |
| P3 | <0.0001 | 0 | 465 | + | 0.0003 | 330 | 48 | - | 0.0003 | 269 | 196 | + | 0.0004 | 443 | 22 | + |
| P4 | <0.0001 | 465 | 0 | + | 0.0006 | 344 | 121 | + | 0.6023 | 157 | 308 | - | 0.0001 | 400 | 51 | + |
| P5 | 0.0001 | 465 | 0 | + | 0.703 | 264 | 201 | - | <0.0004 | 451 | 0 | + | <0.0001 | 453 | 12 | + |
| P8 | 0.6101 | 200 | 265 | - | 0.0001 | 465 | 0 | + | 0.0001 | 279 | 186 | + | 0.0006 | 307 | 158 | + |
| +/=/- | 3/0/1 | | | | 2/0/2 | | | | 3/0/1 | | | | 4/0/0 | | | |

**Table (5): Statistical evaluation utilizing the two-sided Wilcoxon Signed-Rank Test (α=0.05) to identify the best solution for the problems solved in Evaluation 1 (Nvar2)**

| Problems | WOA vs. BSA | | | | WOA vs. FDO | | | | WOA vs. PSO | | | | WOA vs. FF | | | |
|---|---|---|---|---|---|---|---|---|---|---|---|---|---|---|---|---|
| | p-value | T+ | T- | Winner | p-value | T+ | T- | Winner | p-value | T+ | T- | Winner | p-value | T+ | T- | Winner |
| P3 | 0.0004 | 440 | 25 | + | <0.0001 | 61 | 374 | + | 0.0001 | 465 | 0 | + | 0.0001 | 440 | 25 | + |
| P4 | 0.0001 | 444 | 21 | + | 0.5001 | 214 | 0 | = | <0.0001 | 460 | 5 | + | 0.0004 | 455 | 1 | + |
| P5 | 0.5001 | 222 | 243 | - | 0.513 | 185 | 280 | - | 0.5263 | 210 | 255 | - | <0.52001 | 315 | 148 | - |
| P8 | 0.0001 | 433 | 32 | + | 0.0001 | 301 | 164 | + | <0.0001 | 432 | 33 | + | 0.0001 | 355 | 110 | + |
| +/=/1 | 3/0/1 | | | | 2/1/1 | | | | 3/0/1 | | | | 3/0/1 | | | |

**Table (6): Statistical evaluation utilizing the two-sided Wilcoxon Signed-Rank Test (α=0.05) to identify the best solution for the problems solved in Evaluation 1 (Nvar3)**

| Problems | WOA vs. BSA | | | | WOA vs. FDO | | | | WOA vs. PSO | | | | WOA vs. FF | | | |
|---|---|---|---|---|---|---|---|---|---|---|---|---|---|---|---|---|
| | p-value | T+ | T- | Winner | p-value | T+ | T- | Winner | p-value | T+ | T- | Winner | p-value | T+ | T- | Winner |
| P3 | 0.0001 | 440 | 25 | + | 0.5101 | 249 | 216 | - | 0.6101 | 266 | 11 | - | 0.0001 | 425 | 40 | + |
| P4 | 0.5816 | 250 | 210 | = | 0.602 | 211 | 254 | - | 0.786 | 273 | 192 | + | 0.0621 | 464 | 1 | + |
| P5 | <0.0001 | 315 | 0 | + | 0.0001 | 99 | 366 | + | 0.0001 | 465 | 0 | + | 0.0001 | 463 | 2 | + |

| P8 | 0.0001 | 461 | 4 | + | 0.0631 | 462 | 3 | + | 0.0802 | 406 | 0 | + | 0.0761 | 453 | 12 | + |
| +/=/- | 3/1/0 | | | | 2/0/2 | | | | 3/0/1 | | | | 4/0/0 | | | |

On the other hand, every method in Evaluation 2 resolved a subset of optimization problems in order of increasing difficulty. This evaluation resolved two-dimensional optimization issues in R1, R2, and R3—three distinct search spaces. In contrast to Evaluation 1, the number of search spaces impacted the success rates of the algorithms used to minimize the problems since each approach may have removed a different number of problems in each search region. For instance, among 16 optimization tasks, the R1, R2, and R3 techniques removed 11, 9, and 8. The complexity of the solved benchmark functions ranged from low to high as well. For instance, the comparative success rates for identical problems in search space R3 were 6.08 percent and 82.75 percent. A significant difficulty score in search space R3 was also unable to decrease by any methods. In search area R3, P16 had an overall success rate of 62.67 percent and was unsolvable by any method.

The success rate of P7, which was 4.92 percent overall, could not be reduced by any of the algorithms operating in the same search space. This indicates that it is impossible to accurately determine how successful WOA and other algorithms are at reducing optimization problems based on their difficulty ratings. As a result, there is no inherent connection between the complexity of minimization optimization issues and the efficiency of these methods. Like Evaluation 1, Evaluation 2 used the Wilcoxon signed-rank test with a significance level ($\alpha$) of 0.05 to compare WOA against other algorithms. The examined null hypothesis (H0) in this situation is given by Equation (1) for a particular benchmark problem. Tables 7, 8, and 9 show the algorithms that outperformed the other algorithms in Evaluation 2 based on the results of the Wilcoxon signed-rank test.

**Table (7): Statistical evaluation utilizing the two-sided Wilcoxon Signed-Rank Test (α=0.05) to identify the best solution for the problems solved in Evaluation 2 (R1)**

| Problem | WOA vs. BSA | | | | WOA vs. FDO | | | | WOA vs. PSO | | | | WOA vs. FF | | | |
|---|---|---|---|---|---|---|---|---|---|---|---|---|---|---|---|---|
| | p-value | T+ | T- | Winner | p-value | T+ | T- | Winner | p-value | T+ | T- | Winner | p-value | T+ | T- | Winner |
| P2 | 0.0001 | 461 | 4 | + | 0.5701 | 106 | 359 | - | 0.0001 | 459 | 6 | + | 0.0001 | 460 | 5 | + |
| P4 | 0.5138 | 174 | 291 | - | 0.0001 | 445 | 20 | + | 0.541 | 167 | 298 | - | 0.5204 | 156 | 309 | - |
| P5 | 0.0001 | 410 | 55 | + | 0.0001 | 423 | 42 | + | 0.0001 | 461 | 4 | + | 0.0003 | 451 | 14 | + |
| P6 | 0.0001 | 444 | 21 | + | 0.5273 | 351 | 114 | - | 0.0001 | 464 | 1 | + | 0.6198 | 249 | 216 | - |
| P8 | 0.5831 | 364 | 14 | - | 0.691 | 203 | 262 | - | 0.0711 | 324 | 141 | + | 0.0001 | 369 | 96 | + |
| P9 | 0.0001 | 350 | 115 | + | <0.5121 | 105 | 360 | - | 0.0001 | 145 | 320 | + | 0.0701 | 422 | 43 | + |
| P11 | <0.0001 | 461 | 4 | + | 0.0801 | 460 | 5 | + | 0.5004 | 369 | 96 | - | 0.0001 | 322 | 143 | + |
| P12 | 0.7024 | 115 | 350 | - | 0.0001 | 465 | 0 | + | <0.0001 | 154 | 311 | + | <0.0001 | 231 | 234 | + |
| P13 | 0.0001 | 425 | 40 | + | 0.551 | 166 | 299 | - | 0.0001 | 464 | 1 | + | 0.0001 | 371 | 94 | + |
| P14 | 0.0593 | 264 | 210 | + | 0.0611 | 461 | 4 | + | 0.0801 | 16 | 284 | + | 0.0902 | 145 | 320 | + |
| P15 | 0.0001 | 463 | 2 | + | 0.0001 | 261 | 204 | + | 0.0001 | 364 | 101 | + | 0.0321 | 458 | 7 | + |
| +/=/- | 9/0/3 | | | | 6/0/5 | | | | 9/0/2 | | | | 9/0/2 | | | |

**Table (8): Statistical evaluation utilizing the two-sided Wilcoxon Signed-Rank Test (α=0.05) to identify the best solution for the problems solved in Evaluation 2 (R2)**

| Problem | WOA vs. BSA | | | | WOA vs. FDO | | | | WOA vs. PSO | | | | WOA vs. FF | | | |
|---|---|---|---|---|---|---|---|---|---|---|---|---|---|---|---|---|
| | p-value | T+ | T- | Winner | p-value | T+ | T- | Winner | p-value | T+ | T- | Winner | p-value | T+ | T- | Winner |
| P2 | 0.0301 | 401 | 34 | + | 0.672 | 220 | 245 | = | <0.0001 | 407 | 11 | + | 0.0001 | 218 | 0 | + |
| P4 | 0.0001 | 423 | 39 | + | 0.0001 | 201 | 133 | + | 0.5001 | 253 | 211 | - | <0.0001 | 225 | 99 | + |
| P8 | 0.5189 | 264 | 61 | - | 0.5501 | 465 | 0 | - | 0.0001 | 225 | 99 | + | 0.5211 | 265 | 143 | - |
| P9 | 0.0201 | 465 | 0 | + | <0.0001 | 147 | 61 | + | 0.5001 | 125 | 211 | = | <0.0001 | 215 | 100 | + |
| P11 | 0.5201 | 146 | 261 | - | 0.521 | 332 | 101 | - | <0.0001 | 241 | 14 | + | 0.0003 | 178 | 217 | + |
| P12 | 0.0001 | 378 | 0 | + | 0.2001 | 201 | 211 | + | 0.641 | 253 | 131 | - | <0.0001 | 274 | 14 | + |
| P13 | 0.6001 | 364 | 119 | - | 0.0001 | 228 | 29 | + | <0.0001 | 208 | 211 | + | 0.0001 | 365 | 0 | + |
| P14 | 0.0501 | 13 | 265 | + | <0.0001 | 235 | 1 | + | <0.0001 | 109 | 201 | + | <0.0001 | 205 | 4 | + |
| P15 | 0.3001 | 265 | 0 | + | 0.6201 | 165 | 249 | - | 0.0004 | 265 | 49 | + | 0.0001 | 465 | 0 | + |
| +/=/- | **6/0/3** | | | | **5/1/3** | | | | **6/1/2** | | | | **8/0/1** | | | |

**Table (9): Statistical evaluation utilizing the two-sided Wilcoxon Signed-Rank Test (α=0.05) to identify the best solution for the problems solved in Evaluation 2 (R3)**

| Problems | WOA vs. BSA | | | | WOA vs. FDO | | | | WOA vs. PSO | | | | WOA vs. FF | | | |
|---|---|---|---|---|---|---|---|---|---|---|---|---|---|---|---|---|
| | p-value | T+ | T- | Winer | p-value | T+ | T- | Winner | p-value | T+ | T- | Winer | p-value | T+ | T- | Winer |
| P2 | 0.5001 | 233 | 220 | = | 0.6001 | 106 | 231 | - | <0.0001 | 250 | 215 | + | <0.0001 | 265 | 97 | + |
| P4 | 0.0001 | 206 | 18 | + | <0.0001 | 151 | 119 | + | 0.6511 | 178 | 177 | - | 0.004 | 151 | 16 | + |
| P8 | 0.6021 | 110 | 261 | - | 0.0003 | 460 | 5 | + | 0.0001 | 225 | 18 | + | 0.5619 | 176 | 201 | - |
| P9 | <0.0001 | 161 | 245 | + | <0.0001 | 301 | 0 | + | <0.0001 | 425 | 0 | + | 0.0001 | 376 | 11 | + |
| P11 | 0.0001 | 232 | 143 | + | 0.2286 | 222 | 163 | + | 0.5130 | 204 | 219 | - | 0.004 | 235 | 12 | + |

| | | | | | | | | | | | | | | | |
|---|---|---|---|---|---|---|---|---|---|---|---|---|---|---|---|
| P12 | <0.0001 | 250 | 99 | + | 0.5201 | 265 | 198 | - | 0.0198 | 28 | 68 | + | <0.0001 | 205 | 200 | + |
| P13 | 0.6001 | 177 | 2218 | - | 0.6301 | 206 | 228 | - | 0.0004 | 151 | 128 | + | 0.0003 | 265 | 212 | + |
| P14 | 0.0213 | 64 | 2212 | + | 0.6098 | 165 | 218 | - | <0.0001 | 231 | 200 | + | 0.0003 | 401 | 0 | + |
| +/=/- | 5/1/2 | | | | 4/0/4 | | | | 6/0/2 | | | | 7/0/1 | | | |

In Tables 4 to 9, "-" denotes situations where the WOA was shown as statistically inferior performance, and the null hypothesis was rejected. '+' denotes situations in which the WOA was statistically superior to the null hypothesis, and the null hypothesis was rejected; '=' denotes situations in which there is no statistically significant difference between the two comparison algorithms when assessing the degree to which problems can be minimized. In pairwise problem-based statistical comparisons of the algorithms, the last rows of Tables 4-9 provide the (+/=/-) summation for statistically significant scenarios indicated by the symbols "+," "=," and "-." When the (+/=/-) data from Evaluations 1 and 2 are examined, it is possible to conclude that WOA outperformed the other comparative algorithms statistically regarding reducing numerical optimization problems.

For most evaluation parts, Evaluations 1 and 2 suggest that WOA is more efficient in resolving numerical optimization problems with varying complexity, variable sizes, and search areas. However, none of the approaches could minimize each of the sixteen benchmark functions successfully. The data presented in Tables 4, 5, and 6 are graphically represented in Figure (2). The figure depicts the findings of Tables 7, 8, and 9 (3).

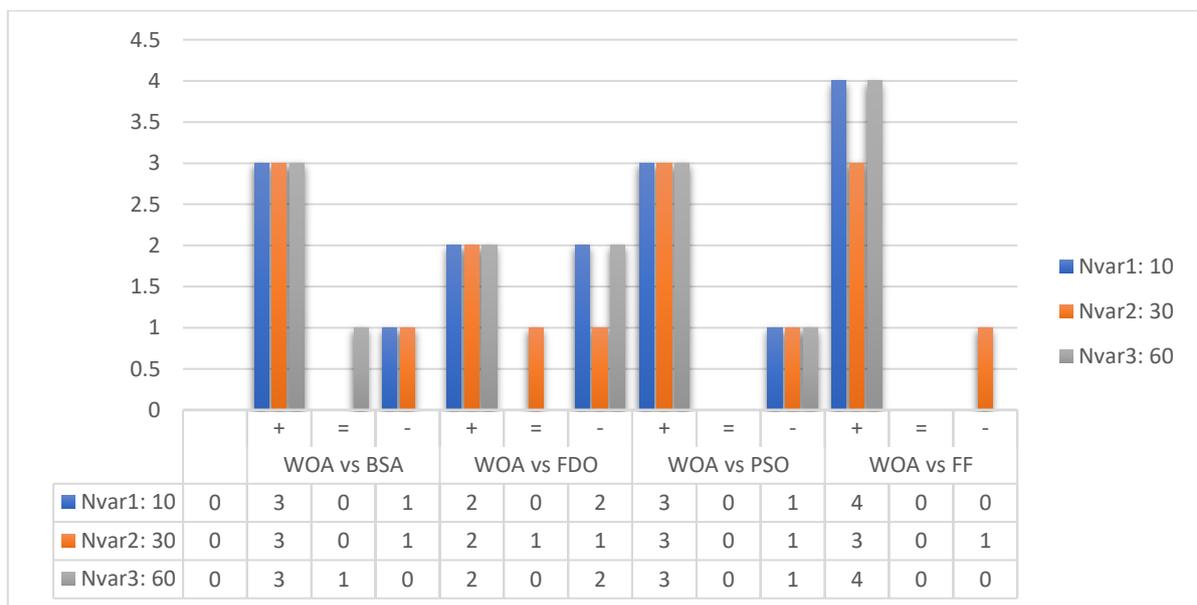

**Figure (2): A graphical form illustration of Evaluation 1**

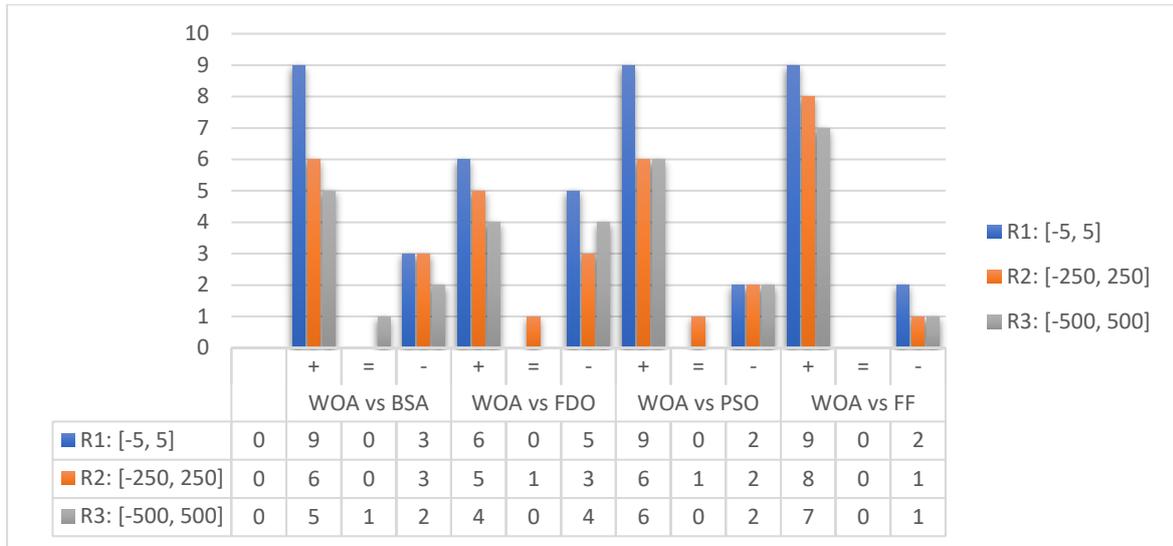

| | + | = | - | + | = | - | + | = | - | + | = | - |
|---|---|---|---|---|---|---|---|---|---|---|---|---|
| | WOA vs BSA | | | WOA vs FDO | | | WOA vs PSO | | | WOA vs FF | | |
| R1: [-5, 5] | 0 | 9 | 0 | 3 | 6 | 0 | 5 | 9 | 0 | 2 | 9 | 0 | 2 |
| R2: [-250, 250] | 0 | 6 | 0 | 3 | 5 | 1 | 3 | 6 | 1 | 2 | 8 | 0 | 1 |
| R3: [-500, 500] | 0 | 5 | 1 | 2 | 4 | 0 | 4 | 6 | 0 | 2 | 7 | 0 | 1 |

**Figure (3): A graphical form illustration of Evaluation 2**

Regarding the first and second evaluations, it was established that none of the algorithms could handle every benchmark issue. Concerning Evaluation 3, the percentage of successful minimization of each of the 16 varied benchmark functions for Nvar1, 2, and 3 with two variable dimensions as the default search space and three different search spaces (R1, R2, and R3). The success and failure rates for decreasing the 16 benchmark functions in Evaluations 1 and 2 are shown in Figures (4) and (5), respectively.

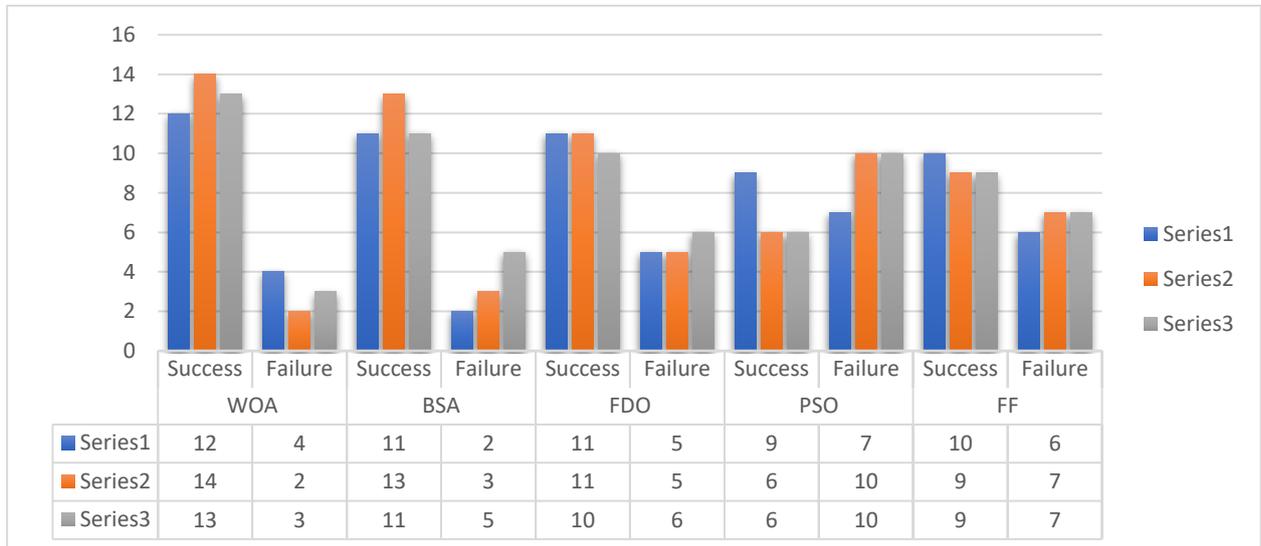

| | Success | Failure | Success | Failure | Success | Failure | Success | Failure | Success | Failure |
|---|---|---|---|---|---|---|---|---|---|---|
| | WOA | | BSA | | FDO | | PSO | | FF | |
| Series1 | 12 | 4 | 11 | 2 | 11 | 5 | 9 | 7 | 10 | 6 |
| Series2 | 14 | 2 | 13 | 3 | 11 | 5 | 6 | 10 | 9 | 7 |
| Series3 | 13 | 3 | 11 | 5 | 10 | 6 | 6 | 10 | 9 | 7 |

**Figure (4): The failure-success ratio for minimizing the problems in Evaluation 1**

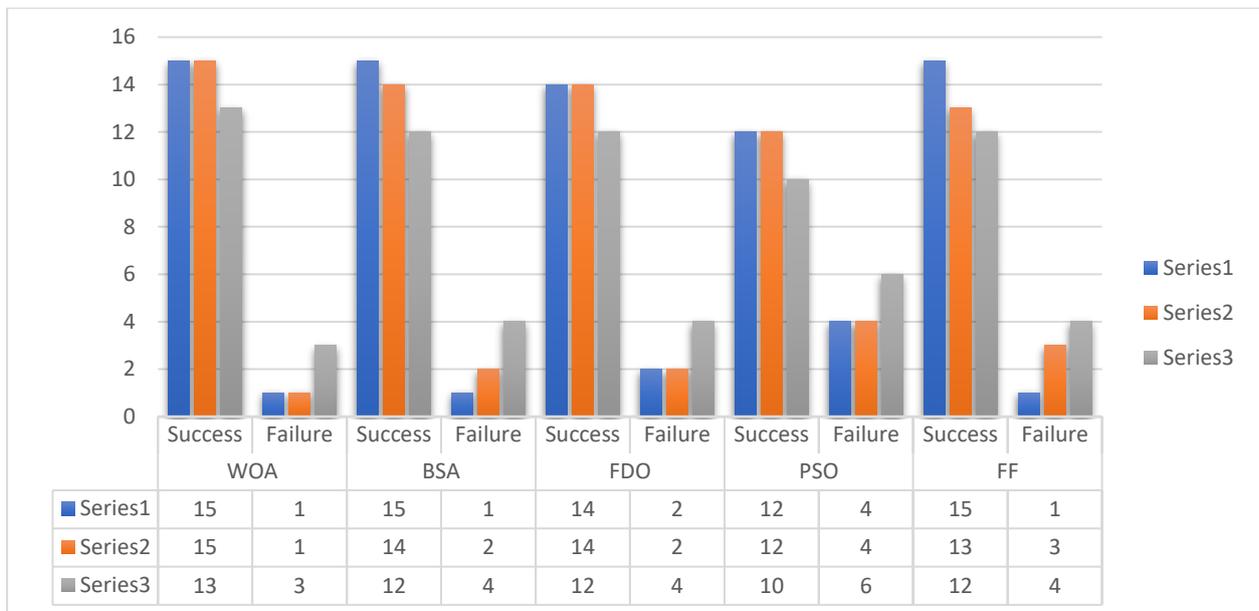

**Figure (5): The failure-success ratio for minimizing the problems in Evaluation 2**

Evaluation 3 determined that WOA was the most effective algorithm for decreasing the highest number of the 16 benchmark functions, while PSO was the least effective technique for attaining the same objective. It also showed that WOA could minimize the most optimization problems; nevertheless, the lowest ratio of successful P1-P16 functions minimization in Nvar1, 2, and 3 FDO uncovered variables that utilized the default search space after that PSO. In contrast, for the variable sizes of two and three different search spaces, WOA and DE had the highest ratio of successful minimization (R1, 2, and 3) of these optimization problems, respectively.

### 4.2. Computational cost

This section analyses the total computational cost of the algorithms (running time/ memory usage) based on the benchmarking problems. The running time and memory usage for the 30 solutions produced by WOA and other algorithms are shown in Table (10). We note that WOA has a faster rate of solving problems. The WOA used less memory than the other methods, too. Surprisingly, FDO needs less running time than WOA for P6 and P12 and uses less memory than WOA for solving P16. At the same time, PSO needs less memory than WOA for P2 and less time for executing P4 than WOA. Referencing FF, it needs less running time than WOA for P11. Also, BSA uses less memory than WOA for solving P12. On average, the WOA executes more quickly and uses less memory than the other optimization techniques.

**Table (10): Average running time and memory usage for the fifteen problems obtained by WOA and its counterpart algorithms**

| Problems | Statistics | WOA | BSA | FDO | PSO | FF | Winner |
|---|---|---|---|---|---|---|---|
| **P1** | **Running time** | 60.465 | 64.625 | 65.692 | 71.318 | 73.444 | WOA |
| | **Memory usage** | 107.610 | 123.882 | 117.958 | 122.546 | 142.770 | WOA |
| **P2** | **Running time** | 87.403 | 108.494 | 88.638 | 131.252 | 97.628 | WOA |
| | **Memory usage** | 168.778 | 190.248 | 187.935 | 150.853 | 187.574 | PSO |
| **P3** | **Running time** | 101.658 | 146.621 | 146.461 | 148.136 | 155.733 | WOA |
| | **Memory usage** | 102.490 | 110.601 | 107.337 | 154.050 | 137.452 | WOA |
| **P4** | **Running time** | 114.398 | 163.755 | 153.260 | 133.616 | 179.608 | PSO |
| | **Memory usage** | 129.673 | 142.252 | 355.075 | 301.916 | 197.261 | WOA |
| **P5** | **Running time** | 96.258 | 211.288 | 175.017 | 114.907 | 157.864 | WOA |
| | **Memory usage** | 139.250 | 168.400 | 161.505 | 196.943 | 155.040 | WOA |
| **P6** | **Running time** | 120.226 | 108.462 | 98.479 | 125.525 | 106.761 | FDO |
| | **Memory usage** | 115.785 | 142.418 | 134.172 | 131.066 | 163.013 | WOA |
| **P7** | **Running time** | 72.416 | 90.503 | 82.868 | 94.466 | 96.738 | WOA |
| | **Memory usage** | 107.937 | 148.801 | 119.474 | 146.541 | 161.103 | WOA |
| **P8** | **Running time** | 154.787 | 322.546 | 251.372 | 187.438 | 373.585 | WOA |
| | **Memory usage** | 218.852 | 265.149 | 304.041 | 308.048 | 329.304 | WOA |
| **P9** | **Running time** | 139.777 | 190.830 | 143.621 | 224.299 | 208.666 | WOA |
| | **Memory usage** | 150.697 | 424.252 | 447.155 | 432.043 | 287.968 | WOA |
| **P10** | **Running time** | 74.609 | 133.127 | 256.108 | 150.900 | 195.263 | WOA |
| | **Memory usage** | 117.771 | 126.280 | 207.945 | 122.908 | 219.695 | WOA |
| **P11** | **Running time** | 190.319 | 330.692 | 297.388 | 312.068 | 176.029 | FF |
| | **Memory usage** | 103.530 | 104.072 | 271.548 | 224.580 | 294.954 | WOA |
| **P12** | **Running time** | 197.392 | 205.910 | 132.140 | 137.415 | 337.746 | FDO |
| | **Memory usage** | 125.762 | 215.917 | 130.117 | 197.576 | 235.339 | WOA |
| **P13** | **Running time** | 125.196 | 145.005 | 170.095 | 133.859 | 173.290 | WOA |
| | **Memory usage** | 96.748 | 133.607 | 129.430 | 245.086 | 381.938 | WOA |
| **P14** | **Running time** | 206.919 | 222.166 | 295.787 | 232.808 | 230.426 | WOA |
| | **Memory usage** | 123.038 | 109.867 | 155.162 | 187.752 | 159.081 | BSA |
| **P15** | **Running time** | 213.554 | 246.275 | 223.216 | 267.662 | 258.846 | WOA |
| | **Memory usage** | 117.521 | 184.633 | 178.929 | 179.451 | 177.233 | WOA |
| **P16** | **Running time** | 116.488 | 181.630 | 173.427 | 137.377 | 151.673 | WOA |

| | Memory usage | 141.698 | 145.050 | 111.549 | 273.085 | 280.214 | FDO |

## 4.2. Convergence analysis

The convergence of WOA is assessed using sixteen benchmark issues (P1–P16). WOA has been independently performed using its comparing algorithms, and the best result has been shown after each repetition. Each algorithm has run 1000 times with a default search space of 30 and a population size of 30. As a result, Figure (5) shows how quickly WOA and its rival algorithms converge. As observed, WOA improved the convergence compared to its comparison methods. Overall, based on the findings from the earlier sections, WOA performs better in exploitation and accelerating convergence. The advantage of WOA over other algorithms is its capacity to avoid local optima and get an optimum global result.

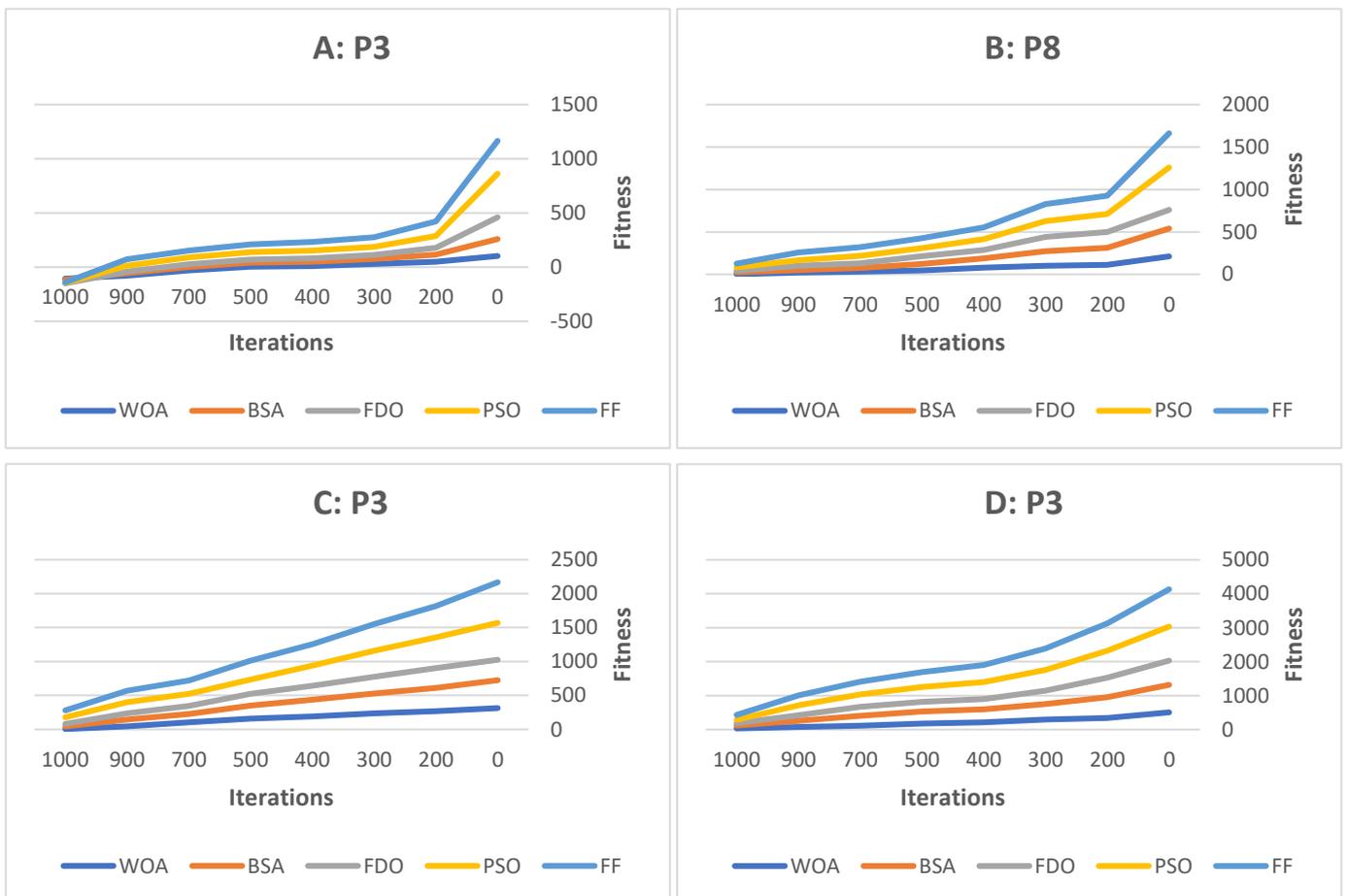

**Figure (5):** Convergence analysis curves of the WOA and its counterpart algorithms on four representative problems; (A) for P3, (B) for P8, (C) for P13 and (D) for P16.

## 5. Summary

In this chapter, the performance of WOA was compared with its analogous algorithms for various real-world problems. The assessment results revealed that WOA is statistically better than competing approaches for reducing various numerical optimization problems without being unduly sensitive to issue dimensions. Although the statistically superior performance was shown between WOA and other algorithms in this study's evaluations, this does not suggest that WOA can be used for all optimization problems regarding hardness score, search space, and dimension problems. In this analysis, WOA could not minimize the sixteen benchmark functions utilized in this experiment. This failure is most likely the result of four factors: variable issue difficulty ratings, search spaces, and the number of dimensions with problem cohorts. The success and failure rates of WOA in resolving various problems with varying difficulty ratings, problem dimensions, and search areas cannot be precisely quantified. Consequently, this experiment illustrates that WOA is somewhat sensitive to the problem's difficulty score, problem dimensions and type, and search space. In addition, WOA was performing better on the sixteen problems in terms of convergence speed, running time, and memory utilization.

problems," *Neural Comput. Appl.*, pp. 1–20, 2020.

[8]  M. Jamil and X.-S. Yang, "A literature survey of benchmark functions for global optimization problems," *arXiv Prepr. arXiv1308.4008*, 2013.

[9]  M. M. Ali, C. Khompatraporn, and Z. B. Zabinsky, "A numerical evaluation of several stochastic algorithms on selected continuous global optimization test problems," *J. Glob. Optim.*, vol. 31, no. 4, pp. 635–672, 2005.

[10] "Global Optimization Benchmarks and AMPGO." [Online]. Available: http://infinity77.net/global_optimization. [Accessed: 24-Nov-2018].

[11] P. Civicioglu, "Artificial cooperative search algorithm for numerical optimization problems," *Inf. Sci. (Ny).*, vol. 229, pp. 58–76, 2013.

[12] B. A. Hassan and T. A. Rashid, "Operational framework for recent advances in backtracking search optimization algorithm: A systematic review and performance evaluation," *Appl. Math. Comput.*, p. 124919, 2019.

[13] J. Derrac, S. García, D. Molina, and F. Herrera, "A practical tutorial on the use of nonparametric statistical tests as a methodology for comparing evolutionary and swarm intelligence algorithms," *Swarm Evol. Comput.*, vol. 1, no. 1, pp. 3–18, 2011.
21